\def\year{2020}\relax
\documentclass[letterpaper]{article} 
\usepackage{aaai20}  
\usepackage{times}  
\usepackage{helvet} 
\usepackage{courier}  
\usepackage[hyphens]{url}  
\usepackage{graphicx} 
\usepackage{graphicx}  
\usepackage{amsmath, amssymb}
\usepackage{multirow}
\title{A Statistical Defense Approach for Detecting Adversarial Examples}
\author{Alessandro Cennamo,\textsuperscript{\rm 1,2} Ido Freeman,\textsuperscript{\rm 1,2} Anton Kummert\textsuperscript{\rm 2}\\ 
\textsuperscript{\rm 1}Aptiv Services Deutschland GmbH, Wuppertal, 42199, Germany\\ 
\textsuperscript{\rm 2}University of Wuppertal (BUW), Wuppertal, 42199, Germany\\
\{alessandro.cennamo, ido.freeman\}@aptiv.com, kummert@uni-wuppertal.de 
}
 
\begin{document}

\maketitle

\begin{abstract}
Adversarial examples are maliciously modified inputs created to fool deep neural networks (DNN). The discovery of such inputs presents a major issue to the expansion of DNN-based solutions. Many researchers have already contributed to the topic, providing both cutting edge-attack techniques and various defensive strategies. In this work, we focus on the development of a system capable of detecting adversarial samples by exploiting statistical information from the training-set. Our detector computes several distorted replicas of the test input, then collects the classifier's prediction vectors to build a meaningful signature for the detection task. Then, the signature is projected onto the class-specific statistic vector to infer the input's nature. The classification output of the original input is used to select the class-statistic vector. We show that our method reliably detects malicious inputs, outperforming state-of-the-art approaches in various settings, while being complementary to other defensive solutions.
\end{abstract}

\noindent The past decade has witnessed the definitive recognition of Deep Learning (DL) algorithms as an efficient tool for tackling a broad variety of tasks, notably image classification. This machine learning technique has progressively found application in diverse areas ranging from medical diagnosis \cite{b10}, through self-driving vehicles \cite{b2} to virtual assistants \cite{b11}. Unfortunately, recent studies have proven that such deep neural networks (DNN) are vulnerable to adversarial examples \cite{b3,b4,b5}: malicious inputs designed to fool a DNN model by making imperceptible modifications to pixel intensities. 

This undermines the classification stability and poses a great issue to the deployment of DNN-based systems, especially in safety-critical applications. 
An attacker, indeed, by altering a traffic sign, might force a fully autonomous vehicle to behave unexpectedly, putting human lives in danger.
Hence, finding a way to identify such adversarial samples has become of paramount importance for twofold reasons: make DL solutions more trustworthy and secure and enable, at the same time, their massive deployment.

Herein, we present a defensive solution which does not affect the performance of the classification system nor constrain the training procedure.
The design of an adversarial detector represents one of the most promising solutions in this direction: 
it allows to predict the nature of an input -- malicious or legitimate -- and decide whether to reject or maintain the outcome of the defended classifier.
In addition, this type of defensive strategy can be considered as an external, separate module, which could be turned on only when the environment is prone to malicious instances.

We propose to combine the robustness of DNNs to greedy image distortions with statistical information from the training-set to build an effective detection framework. We show that distorted adversarial examples have different underlying statistics than their legitimate counterparts. This allows us to build some orthogonality between malicious samples and statistics from the legitimate training-set. Finally, we take advantage of this property to reliably detect them.

Our work provides the following contributions:

\begin{itemize}
	\item We develop a low-cost, effective adversarial detector able to reliably recognize and discard malicious inputs in both white and black-box settings.
	\item We provide a proof of the compatibility property with other defensive technique, such as adversarial training.
	\item We devise a brand new distortion, enabling the adversarial detection task.
\end{itemize}

\section{Background and Related Works}\label{Bg}

Over the years, the development of numerous machine learning (ML) frameworks along with the availability of vast amounts of data have made the task of training DNNs quite simple. Despite this, researchers have only scratched the surface of such networks, which are still treated as black-box entities. The possibility to unveil hidden properties and behaviors of DNNs has started a real arms-race. Since the discovery of adversarial examples, an ever-increasing number of researchers attempted to design new, more powerful attacks capable of defeating the current state-of-the-art defense. In this section we describe some of the most important attacks and defensive strategies present in the literature.

\subsection{Adversarial Attacks}\label{AA}

The development of an attack method requires profound knowledge of intrinsic properties of DNN systems. Adversarial examples are, indeed, the result of an optimization algorithm over the input domain and the possibility to fine tuning such optimization procedure opens up to a variety of different scenarios.

Formally, it is possible to define the task of generating a malicious input \textit{$x'$} out of a legitimate input \textit{$x$} as the act of finding the perturbation $\eta$ such that,
\begin{equation}
x' = x + \eta \label{eq_adv_ex}
\end{equation}
leads to a different classification result.
The perturbation $\eta$ is generally the result of an optimization algorithm.

The Fast Gradient Sign Method (FGSM) attack was originally designed to speed-up the generation process \cite{b1}. It does so by performing a single step along the direction of the gradients w.r.t. the loss function calculated on a generic clean input. Equation \eqref{eqFGSM} shows the computation of the perturbation, where  $l$ is the ground-truth label associated with input  $x$ and $L(x, l)$ represents the loss function. The magnitude of the step is the same for every pixel and is controlled by the step-size parameter $\epsilon$.
\begin{equation}
\eta = \epsilon \cdot sign(\nabla_x L (x, l)) \label{eqFGSM}
\end{equation}
FGSM is one of the earlier attack techniques ever developed. It is very fast compared to other methods but produces more aggressive perturbations, i.e., the malicious inputs are mostly visible to the human eye. However, it is interesting to notice how a very greedy technique can effectively undermine the stability of a neural network.
This is also confirmed by the manifold variants which succeeded FGSM, attempting to improve the efficiency of the method \cite{b5,b6,b7,b8,b9}.

Other authors tried to refine the algorithm to achieve the smallest adversarial perturbation (in terms of magnitude). Moosavi-Dezfooli et al. \cite{bDF} believed that the geometrical representation of DNN models represents a powerful tool for understanding the nature of such malicious samples. They leveraged the hyperplane theory to represent the decision boundaries imposed by a network in the high-dimensional input domain. Finally, they created DeepFool (DF), an iterative procedure to achieve the smallest perturbation moving a natural input \textit{just beyond} its closest decision boundary. 
Compared with FGSM, it produces much finer perturbation masks, at the expense of higher computational time. Moreover, since it pushes the input \textit{just beyond} the closest decision boundary, it does not allow to accurately control the effectiveness of the attack -- sometimes it results in uncertain malicious samples because of their close proximity to the decision boundary.

Carlini and Wagner \cite{bC_W} presented an effective attack technique to be used as benchmark against new attacks/defenses. They mathematically approached the adversarial generation process, providing new ways to address the optimization problem. The resulting attack method is an iterative process named after its inventors Carlini\&Wagner (C\&W). Compared to DF it has the advantage of providing the attacker with several hyper-parameters to tune in order to achieve adversarial samples with a broad variety of properties.
C\&W is actually considered the state-of-the-art attack techniques, as confirmed by the broad variety of works present in literature \cite{bFS,b15,b16}.

Finally, it is worth introducing the definition of two common attack settings: white-box and black-box. The former accounts for an attacker which has full access to the classifier, including the backpropagation signal, while the latter assumes that the attacker does not have access to the attacked model.

\subsection{Defensive Strategies}\label{DS}
The ability of reliably defend against adversarial inputs is of paramount importance to enable massive deployment of DNN-based solutions, especially in sensitive areas. Over the years, researchers have developed various approaches which can be divided in two main families: \textit{proactive} and \textit{reactive}. Strategies of the former type aim at making the defended network more robust during the design and/or training process whilst approaches from the latter family are deployed after the classifier has already been trained, i.e. without tempering the design/training process.

Some researchers believe that learning from adversarial examples would increase DNNs robustness without significantly harming the performance. Goodfellow et al. \cite{b1} proposed to introduce a factor accounting for adversarial examples in the loss function used during training. Adversarial training of a network is a proactive solution: it requires to generate malicious inputs live -- for the current learned network configuration -- and uses the loss produced to perform the next optimization step. In this way, one can increase the model's generalization power to adversarially manipulated inputs.
This technique has proven to successfully increase robustness at the expense of increased training time. In addition, another desired property conferred by adversarial training is the better generalization power, as reported in \cite{b1} and \cite{b17}.
\begin{figure*}[t!]
	\centering\includegraphics[width=1.7\columnwidth,keepaspectratio]{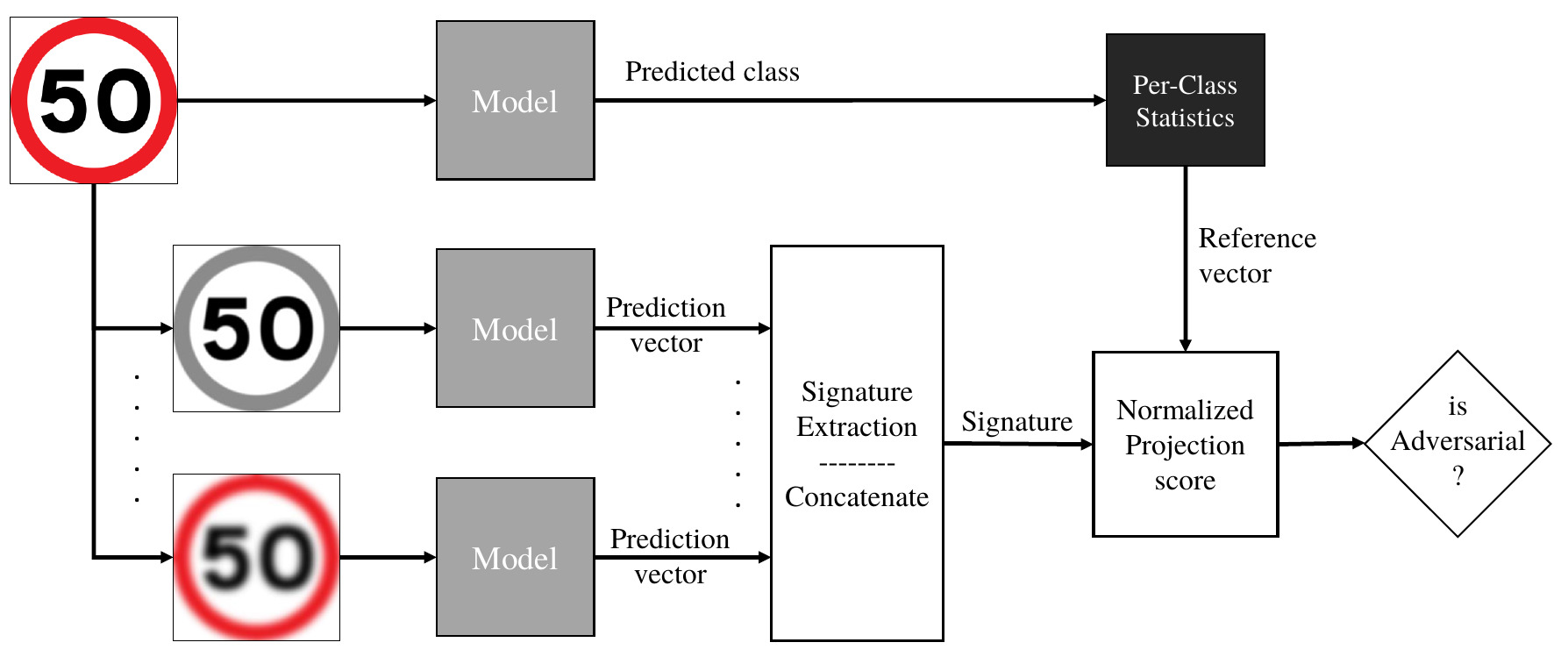}
	\caption{Framework of the proposed detection scheme. Several distorted replicas of the original input are computed. The prediction vectors on every input replica are collected to build a signature. The classification output on the original input is used to select the per-class first-order statistic. The normalized projection score is thresholded to infer the input nature.}
	\label{fig_scheme}
\end{figure*}

Xu et al. \cite{bFS} instead, realized that it is possible to lower the degree of freedom available to an attacker by reducing the input space dimensionality. They proposed to detect malicious samples by using distortions to squeeze the pixel domain.
The resulting system, named Feature Squeezing (FS), acts as a reactive defense: when the classifier receives a generic input, it computes the squeezed version and compares the classifier outcomes of both original and squeezed input. Then, a thresholding operation decides whether the input is adversarial or legitimate.
FS possesses the advantages of being a low-cost attack-agnostic technique, which affects neither the model architecture nor the training procedure. On the other hand, it requires to query the classifier multiple times, thus resulting in higher inference time.

Liao et al.  \cite{b18} considered the possibility to reconstruct the original, legitimate input from an adversarial one. 
They proposed to clean the classifier's input by means of a data-driven pre-processing entity.
The devised denoiser was trained exploiting information from higher levels of the defended network, hence named High-level representation Guided Denoiser (HGD). The authors showed that their proposal efficiently solves the adversarial issue with good transferability properties. Unfortunately, when deployed along with the classifier, it creates a new DL system which is very prone to be attacked in white-box settings. Moreover it is not attack agnostic since it requires adversarial samples to train the Denoiser.


\section{Our Method}\label{Meth}

In this section, we provide a description of our defense approach. We believe that the deployment of an adversarial examples detector represents the most interesting solution to defend DNN models. Indeed, this external module does not affect the classifier's performance and can be activated only when required. Therefore, we designed a detector that exploits the robustness of DNN models to greedy distortions.

As in \cite{bFS}, we first noticed that after being distorted, adversarial examples manifest different statistics than legitimate ones.
Since it is well-known that DNNs are invariant to several distortions, we thought that it could be possible to exploit such distortions to build a meaningful signature for the detection task.
In our work we propose to concatenate the probability vectors outputted by the classifier when queried with the distorted replicas of the received input.

The extracted signature is then compared with a reference vector in order to assess the input nature. We decided to use class-representative vectors: in particular, 
per-class first order statistics, so that legitimate samples belonging to a given class must exhibit statistics similar to the class-representative vector. The reference vectors are computed on the same set used to train the network: the detector extracts the signature vector for every training sample and then averages across samples from the same class. Equation \eqref{eq1ordstat} contains the formal definition of the described procedure, where $f(x)$ is the function learned by the network, $\Psi = \{\psi_i: i = 1, \dots, m\}$ the set of distortions used, $g(\cdot,\Psi)$ the function building the signature, $\lvert \cdot \rvert$ the dimensionality operator returning the number of samples in the set, $n$ and $m$ the total number of classes and distortions, respectively.
\begin{equation}
\mu_j = \frac{1}{\lvert X_j \rvert} \sum_{x \in X_j} g\left(f(x), \Psi \right) \quad \quad \forall\ j = 1, \dots, n
\label{eq1ordstat}
\end{equation}
We use the class predicted on the original input to select the class-specific vector. Fig. \ref{fig_scheme} shows the framework of the detector described.

The scheme described so far forced adversarial examples to produce signatures which are \textit{orthogonal} to the corresponding reference vector (for proper choice of the distortions).
This discovery led us to choose the \textit{normalized projection score} (a.k.a. cosine similarity) as metric in the comparison, 
hence quantifying the similarity among signature and class-representative statistics.
Equation \eqref{eqproj} formally defines the computation, where $j = \arg\max f(x)$ is the predicted class for the original input, $\lVert \cdot \rVert_2$ is the $L_2$--norm and $\gamma(x) \overset{\vartriangle}{=} g\left(f(x), \Psi \right)$  for brevity -- the function building the signature in Eq. \eqref{eq1ordstat}. Technically, 
it computes the cosine of the angle between the two vectors under comparison.
\begin{equation}
proj(x) = \frac{\gamma(x)^T \mu_j}{\lVert \gamma(x) \rVert_2 \cdot \lVert \mu_j \rVert_2}
\label{eqproj}
\end{equation}
The higher the score is, the more aligned the two vectors are, the more likely is the input to be legitimate. 

\begin{figure}[t!]
	\centering\includegraphics[width=\columnwidth,keepaspectratio]{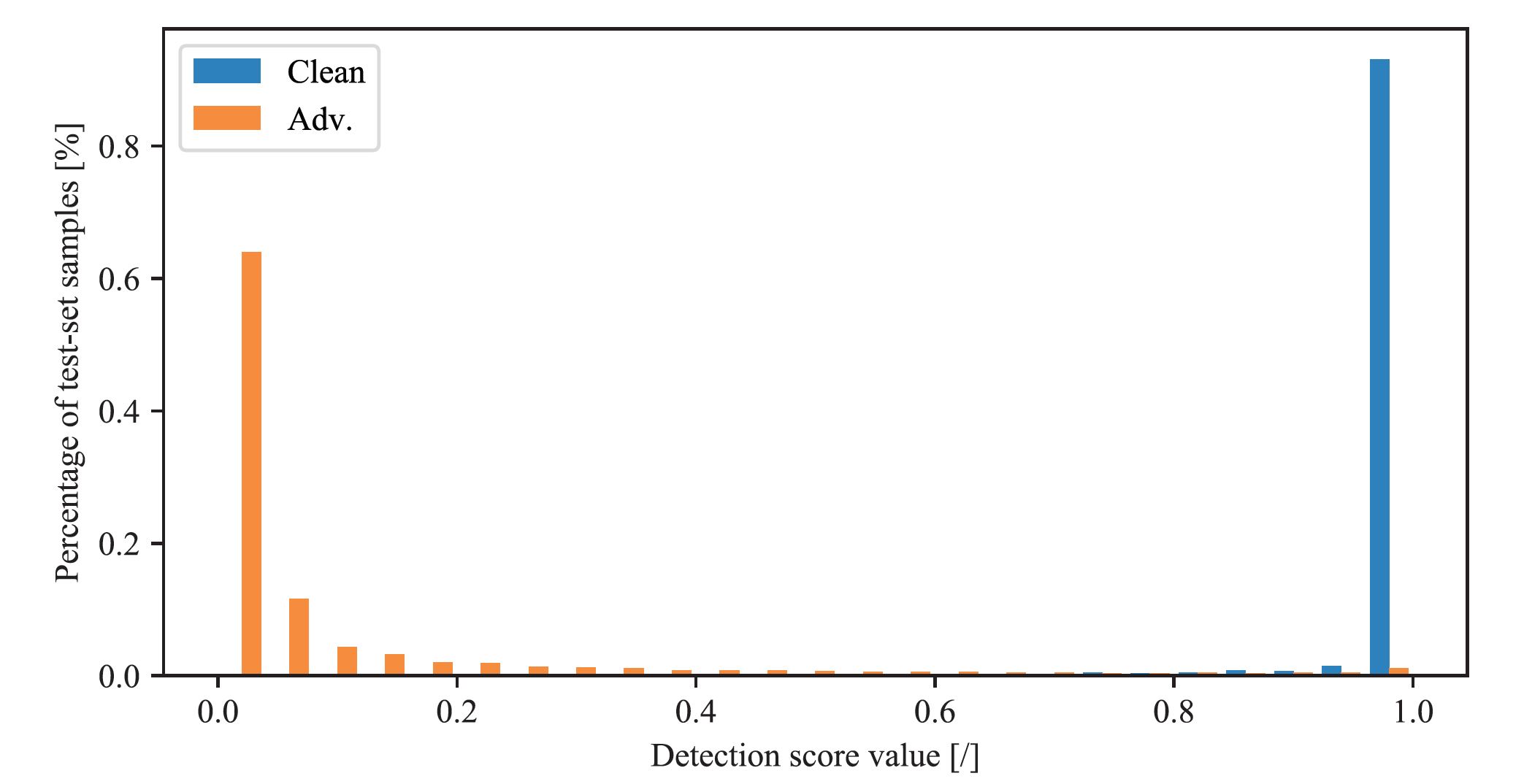}
	\caption{Histogram plotting the distribution the score extracted by the proposed detector for legitimate (blue) and adversarial (orange) samples. Inputs were sampled from the GTSRB test-set and perturbed using the C\&W method. The detector used a single distortion: median filtering. The separation among the two different class is very neat.}
	\label{fig_hist}
\end{figure}


\begin{table*}[t]
	\def\arraystretch{1.3}
	\scriptsize
	\centering
	\begin{tabular}{|c||c||c|*{6}{c}|*{3}{c}|}
		\cline{2-12} 
		
		\multicolumn{1}{c}{} & \multicolumn{1}{|c||}{} & \textbf{Legitimate} & \multicolumn{6}{c|}{\textbf{White-box}} &  
		\multicolumn{3}{c|}{\textbf{Black-box}} \\
		
		\cline{4-12} 
		
		\multicolumn{1}{c}{} &  \multicolumn{1}{|c||}{\textbf{Metrics}} & \textbf{Samples} & \textbf{C\&W} & \textbf{C\&W5} & \textbf{C\&W9} & \textbf{DF} & \textbf{FGSM1} & \textbf{FGSM4} & \textbf{C\&W} & \textbf{DF} & \textbf{FGSM4}\\
		
		\cline{2-12} \noalign{\smallskip} \hline
		
		
		\multirow{3}*{\rotatebox[origin=c]{90}{\parbox[c]{1.78cm}{\centering \textbf{CIFAR10}}}}&  \textbf{Accuracy (victim)}  & 76.83  & 13.32 & 13.21 & 13.21 & 13.97 & 21.08 & 12.66 & 75.68 & 75.63 & 38.50 \\
		
		\cline{2-12} 
		
		& \textbf{Avg. Prediction}  & \multirow{2}*{84.14} &  \multirow{2}*{44.86} &  \multirow{2}*{55.79} & \multirow{2}*{63.88} & \multirow{2}*{44.58} & \multirow{2}*{74.83} & \multirow{2}*{78.87} & \multirow{2}*{59.30} & \multirow{2}*{59.31} & \multirow{2}*{65.39} \\

		& \textbf{Confidence (victim)} & & & & & & & & & &\\
		\cline{2-12} 
		
		& \textbf{Accuracy (substitute)}  & 79.11 & --- & --- & --- & --- & --- & --- & 11.80 & 12.84 & 11.89\\
		
		\hline \noalign{\smallskip} \hline
		
		
		\multirow{3}*{\rotatebox[origin=c]{90}{\parbox[c]{1.83cm}{\centering \textbf{GTSRB}}}}& \textbf{Accuracy (victim)}  & 92.88  & 3.66 & 3.66 & 3.64 & 3.52 & 42.19 & 20.58 & 92.71 & 92.66 & 78.97 \\
		
		\cline{2-12} 
		
		& \textbf{Avg. Prediction}  & \multirow{2}*{96.88} &  \multirow{2}*{51.55} &  \multirow{2}*{63.17} & \multirow{2}*{71.41} & \multirow{2}*{48.45} & \multirow{2}*{83.90} & \multirow{2}*{91.58} & \multirow{2}*{63.58} & \multirow{2}*{63.35} & \multirow{2}*{66.95} \\

		& \textbf{Confidence (victim)} & & & & & & & & & &\\
		\cline{2-12} 
		
		& \textbf{Accuracy (substitute)}  & 96.72 & --- & --- & --- & --- & --- & --- & 1.72 & 2.03 & 20.47\\
		
		\hline

	\end{tabular}
		\caption{Performance of the tested models against every test-set. Accuracies and average prediction confidences are reported (in percentage, \%) for both legitimate and adversarial test-sets when no adversarial defense is deployed.} 
		\label{tab_accs}
\end{table*} 
Fig. \ref{fig_hist} contains the histogram produced by our detector on the GTSRB dataset \cite{bGTSRB}. Legitimate samples get scores close to $1$, whilst adversarial ones are almost all close to be orthogonal. It proves the validity of our detection scheme.

Before delving into the results section, it is worth discussing the distortions used in our work. In order to enable fair comparison with previous works, we borrowed some of the distortions from \cite{bFS}, namely \textit{median filtering} and \textit{bit-depth reduction}. The former consists of a simple spatial smoothing operation, replacing the pixel intensities with the median value of their neighbors, whilst the latter re-quantizes the intensity values with a reduced number of bits -- common camera sensors acquire images with 8-bit resolution pixel intensities per channel. In addition, it was devised a new distortion which converts RGB images to the corresponding gray-scale version. In order to maintain the model's information flow, the distortion rearranges the input as expected by the classifier: several copies of the single channel, gray-scale image are stacked on top of each other to generate a 3 channel input.\footnote{The distortion was tested only on the RGB color-space representations.}

We will refer to the described distortion as \textit{gray-scale}.

\section{Results}\label{Res}

In this section we describe the experimental setup used and report the results of the experiments carried out.

Our proposal was tested against two different datasets: GTSRB \cite{bGTSRB} and CIFAR10 \cite{bCIFAR10}. The former contains a collection of 43 traffic signs organized in 40k training and 13k test samples circa. The samples come in an RGB format with different shapes. Every image is resized to 46x46x3 before feeding the classifier. The latter consists of 10 different classes and provides 50k training and 10k test samples. The images come in an RGB format of shape 32x32x3.

A critical point during the evaluation of any defensive strategy is the selection of the attack methods. 
In this work we decided to focus on three different attacks for different reasons:
\begin{itemize}
	\item DF represents a very strong technique, able to generate adversarial inputs with imperceptible perturbations.
	\item C\&W is the current state-of-the-art attack for benchmarking new defenses; beyond producing invisible perturbations, it allows the attacker to control the effectiveness of the attack through the customizable parameter $\kappa$.
	\item FGSM produces very aggressive perturbations, sometimes visible to humans; however, it is of interest to study the detector behavior against such kind of malicious inputs.
\end{itemize}
The adversarial sets were generated in both white and black-box attack settings (crafted from a substitute model trained on the same dataset).
Table \ref{tab_accs} reports the performance of the trained models on both legitimate and adversarial test-sets. The number beside C\&W specifies the value of the parameter $\kappa$ -- e.g. C\&W9 has  $\kappa = 0.9$ -- while the number beside FGSM stresses out the magnitude of the step size parameter $\epsilon$ -- e.g. FGSM1 has $\epsilon = 0.01$. 
The substitute models were devised to have more capacity than the relative victim.

In terms of results, we compare our method with Feature Squeezing (FS) which represents the closest state-of-the-art solution \cite{bFS}. Indeed, it shares with our method the properties of being a low-cost, model-free technique, 
thus enabling a fair comparison. 

For each experiment, it was chosen to extract the detection ROC curves and provide the comparison as AUC score in order to be independent to the threshold level. The ROC curves were computed pairing the attack-set with an equal number of correctly predicted samples from the original test-set.

Unless otherwise stated, every detector was tested using both the distortions found in literature, namely \textit{median filtering} and \textit{bit-depth reduction}.

\subsection{White-box Results}\label{WBR}

\begin{figure}[b!]
	\centering\includegraphics[width=.95\columnwidth,keepaspectratio]{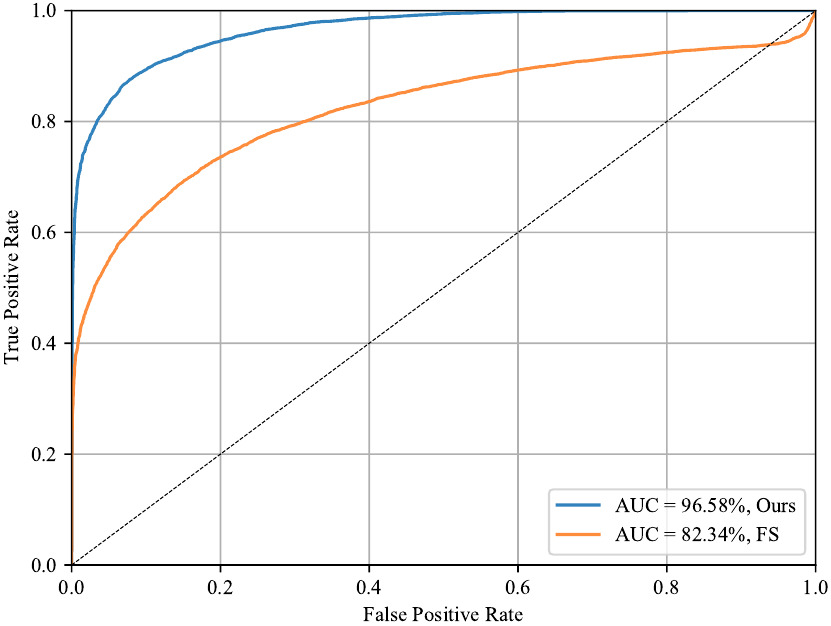}
	\caption{ROC curves of our (blue) and FS (orange) detector against the C\&W attack on the CIFAR10 test-set. 
		Both the detectors use median filtering and reduction of the bit-depth as distortions. 
	}
	\label{fig_ROC}
\end{figure}

Fig. \ref{fig_ROC} shows the ROC curves for both FS and our detectors on the CIFAR10 dataset. The improvement introduced by our method is very clear, passing closer to the top-left corner and demonstrating  a more convergent behavior.

Table \ref{tab_auc}, reports the AUC scores produced for all the white-box attack-sets. 
It confirms that our detector outperforms FS on the strongest attack-sets, i.e. C\&W and DF, whilst FS is able to perform as well as our method (or better) when tested against the FGSM attacks. However, the latter aspect does not represent a serious issue, as shown in section \ref{CwAT}.

\begin{table}[b!]
	\def\arraystretch{.9}
	\centering
	\scriptsize
	\begin{tabular}{|c||c||*{6}{c}|}
		\cline{2-8} 
		
		\multicolumn{1}{c}{} \rule{0cm}{7pt} & \multicolumn{1}{|c||}{\textbf{Meth.}} & \textbf{CW} & \textbf{CW5} & \textbf{CW9} & \textbf{DF} & \textbf{FGSM1} &  \textbf{FGSM4} \\
		
		\cline{2-8} \noalign{\smallskip} \hline
		
		\multirow{4}*{\rotatebox[origin=c]{90}{\parbox[c]{1.1cm}{\centering \textbf{CIFAR10}}}}& \multirow{2}*{\textbf{FS}}  &  \multirow{2}*{82.34}  &  \multirow{2}*{86.62}  &  \multirow{2}*{89.45}  &  \multirow{2}*{82.21}  &  \multirow{2}*{\textbf{88.89}}  &  \multirow{2}*{\textbf{74.90}}  \\
		
		&  & & & & & & \\
		
		\cline{2-8} 
		
		& \multirow{2}*{\textbf{Ours}}  &  \multirow{2}*{\textbf{96.58}}  &  \multirow{2}*{\textbf{95.57}}  &  \multirow{2}*{\textbf{95.31}}  &  \multirow{2}*{\textbf{97.00}}  &  \multirow{2}*{84.17}  &  \multirow{2}*{67.52}  \\
		
		&  & & & & & & \\
		
		\hline \noalign{\smallskip} \hline
		
		
		\multirow{3}*{\rotatebox[origin=c]{90}{\parbox[c]{1.05cm}{\centering \textbf{GTSRB}}}}& \multirow{2}*{\textbf{FS}}  &  \multirow{2}*{96.55}  &  \multirow{2}*{97.76}  &  \multirow{2}*{98.39}  &  \multirow{2}*{96.16}  &  \multirow{2}*{\textbf{93.74}}  &  \multirow{2}*{73.60}  \\
		
		&  & & & & & & \\
		
		\cline{2-8} 
		
		& \multirow{2}*{\textbf{Ours}}  &  \multirow{2}*{\textbf{99.77}}  &  \multirow{2}*{\textbf{99.69}}  &  \multirow{2}*{\textbf{99.61}}  &  \multirow{2}*{\textbf{99.62}}  &  \multirow{2}*{92.54}  &  \multirow{2}*{\textbf{74.08}}  \\
		
		&  & & & & & & \\
		
		\hline
	\end{tabular} 
	\caption{AUC scores of FS and our detector against every attack-set. All results are reported in percentage (\%).}
	\label{tab_auc}
\end{table}

Furthermore, bearing table \ref{tab_accs} in mind
, it is interesting to see how the effectiveness of FS significantly drops when the classifier to defend comes with more uncertainty (i.e. CIFAR10 model), thus suggesting that our detector is more robust to changes in performance of the defended model.

\subsection{Compatibility with Adversarial Training}\label{CwAT}

The previous subsection showed that our detector is highly effective against attacks which produce fine perturbations. Simultaneously, it showed a lack of efficiency in detecting more aggressive attacks. Luckily, the literature is full of solutions capable to tackle such malicious inputs.

In this section we compare the performance of FS and our detector when deployed along with another defensive strategy: adversarial training \cite{b1}. In more details, the classifier to defend was fine trained for a few epochs, 
substituting half of the samples in the mini-batch with the corresponding FGSM4 adversarial version against the current network configuration. Every adversarial attack-set was then computed again on the final, adversarially trained version of the network.

Table \ref{tab_accs_AT} contains the performance of the models with and without adversarial training. As expected, the results are significantly more robust to FGSM attacks, suggesting that the network learned the heavy pattern of the adversarial perturbations. On the other hand, we notice a drawback in the classifier's performance, manifesting a slight drop in accuracy. Fine tuning the adversarial training process allows the prevention of this effect.

\begin{table}[b!]
	\def\arraystretch{1.}
	\centering
	\scriptsize
	\begin{tabular}{|c||c||c|*{4}{c}|}
		\cline{2-7} 
		
		\multicolumn{1}{c}{} \rule{0cm}{7pt}
		&  \multicolumn{1}{|c||}{\textbf{Metric}} & \textbf{Legit.} & \textbf{C\&W} & \textbf{DF} & \textbf{FGSM1} & \textbf{FGSM4} \\
		
		\cline{2-7} \noalign{\smallskip} \hline
		
		
		\multirow{3}*{\rotatebox[origin=c]{90}{\parbox[c]{1.35cm}{\centering \textbf{CIFAR10}}}}& \multirow{1}*{\textbf{Accuracy w/o}}  &  \multirow{2}*{76.83}  &  \multirow{2}*{13.32}  &  \multirow{2}*{13.97}  &  \multirow{2}*{21.08}  &  \multirow{2}*{12.66} \\
		
		& \multirow{1}*{\textbf{adv. training}}  & & & & & \\
		
		\cline{2-7} 
		
		& \multirow{1}*{\textbf{Accuracy w.}}  &  \multirow{2}*{73.58}  &  \multirow{2}*{14.18}  &  \multirow{2}*{14.29}  &  \multirow{2}*{46.65}  &  \multirow{2}*{28.16} \\

		& \multirow{1}*{\textbf{adv. training}} & & & & & \\
		\cline{2-7}

		\hline \noalign{\smallskip} \hline
		
		
		\multirow{3}*{\rotatebox[origin=c]{90}{\parbox[c]{1.4cm}{\centering \textbf{GTSRB}}}}& \multirow{1}*{\textbf{Accuracy w/o}}  &  \multirow{2}*{92.88}  &  \multirow{2}*{3.66}  &  \multirow{2}*{3.52}  &  \multirow{2}*{42.19}  &  \multirow{2}*{20.58} \\
		
		&  \multirow{1}*{\textbf{adv. training}} & & & & & \\
		
		\cline{2-7} 
		
		& \multirow{1}*{\textbf{Accuracy w.}}  &  \multirow{2}*{91.95}  &  \multirow{2}*{4.31}  &  \multirow{2}*{4.22}  &  \multirow{2}*{73.49}  &  \multirow{2}*{31.80} \\

		&  \multirow{1}*{\textbf{adv. training}} & & & & & \\
		\cline{2-7}

		\hline

	\end{tabular}
	\caption{Accuracies of the models before and after adversarial training. All results are reported in percentage (\%).}
	 \label{tab_accs_AT}
\end{table}

\begin{table}[b!]
	\def\arraystretch{1.3}
	\centering
	\scriptsize
	\begin{tabular}{|c||c|c||*{4}{c}|}
		\cline{2-7} 
		
		\multicolumn{1}{c}{} 
		&  \multicolumn{1}{|c|}{\textbf{Method}} & \multicolumn{1}{c||}{\textbf{Adv. Train}} & \textbf{C\&W} & \textbf{DF} & \textbf{FGSM1} & \textbf{FGSM4} \\
		
		\cline{2-7} \noalign{\smallskip} \hline
		
		
		\multirow{3}*{\rotatebox[origin=c]{90}{\parbox[c]{1.45cm}{\centering \textbf{CIFAR10}}}}& \multirow{2}*{\textbf{FS}}  &  \multirow{1}*{\textbf{w/o}}  & 82.34  & 82.21 & \textbf{88.89} & 74.90 \\
		
		\cline{3-7}
		
		&  &\multirow{1}*{\textbf{w.}} & 81.40 & 80.73 & 79.81 & 72.50\\
		
		\cline{2-7} 
		
		&  \multirow{2}*{\textbf{Ours}}   &  \multirow{1}*{\textbf{w/o}}  & 96.58 & \textbf{97.00} & 84.17 & 67.52 \\
		
		\cline{3-7}
		&  & \multirow{1}*{\textbf{w.}} & \textbf{96.65} & 96.73 & 86.48 & \textbf{78.41} \\
		\cline{2-7}

		\hline \noalign{\smallskip} \hline
		
		
		\multirow{3}*{\rotatebox[origin=c]{90}{\parbox[c]{1.5cm}{\centering \textbf{GTSRB}}}}& \multirow{2}*{\textbf{FS}}  &  \multirow{1}*{\textbf{w/o}}  & 96.55  &96.16 & 93.74 & 73.60 \\
		
		\cline{3-7}
		
		&  &\multirow{1}*{\textbf{w.}} & 88.39 & 88.61 & 92.20 & 89.45 \\
		
		\cline{2-7} 
		
		&  \multirow{2}*{\textbf{Ours}}   &  \multirow{1}*{\textbf{w/o}}  & \textbf{99.77} & \textbf{99.62} & 92.54 & 74.08 \\
		
		\cline{3-7}
		&  & \multirow{1}*{\textbf{w.}} & 93.26 & 93.26 & \textbf{95.77} & \textbf{92.81}\\
		\cline{2-7}

		\hline
		
	\end{tabular}
	\caption{AUC scores of FS and our detector without and with adversarial training. All results are in percentage (\%).}
	 \label{tab_auc_AT}
\end{table}

Table \ref{tab_auc_AT} reports the AUC scores produced by both FS and our detector deployed with and without adversarial training. 
We notice that adversarially training the classifier helps the detection task, especially on the FGSM attack-sets. Moreover, the results shows that our approach comes with better compatibility property than FS, benefiting the most from adversarial training. Last but not the least, it is important to consider the slight accuracy drop of the adversarially trained networks during our analysis, because it accounts for poorer results against both C\&W and DF attack-set -- as shown in section \ref{WBR}.

\subsection{Distortions Configuration}\label{Dist}

In this section we provide an analysis of the detector behavior using different configuration of distortions. We tested both FS and our detectors with a single distortion (using those introduced in section \ref{Meth}), two distortions and all three distortions.

The results are presented in table \ref{tab_auc_dist}. It is clear that our detector benefits from the use of several distortions: the configurations which use multiple distortions nearly always obtain the best scores. The same does not hold for FS, achieving its peak performance when the single-distortion settings are adopted. These insights suggest that the detector developed is able to build a meaningful signature for the detection task, hence combining the effect of multiple distortions in a better way than FS does.

Focusing on the single-distortion configurations, we notice the new \textit{gray-scale} distortion to perform nicely, outperforming the \textit{bit-depth reduction} on the FGSM4 attack-set. This proves that it represents a valid distortion, especially when used in multiple-distortions configurations.
Finally, we find that the effectiveness of each distortion is problem-dependent, as certifies the different results achieved on CIFAR10 and GTSRB datasets.

\begin{table}[b!]
	\def\arraystretch{1.3}
	\centering
	\scriptsize
	\begin{tabular}{|c||c||*{6}{c}|}
		\cline{2-8} 
		
		\multicolumn{1}{c}{} 
		&  \multicolumn{1}{|c||}{\textbf{}} &  \multicolumn{2}{c|}{\textbf{C\&W}} & \multicolumn{2}{c|}{\textbf{DF}} & \multicolumn{2}{c|}{\textbf{FGSM4}} \\
		
		\cline{3-8}
		
		\multicolumn{1}{c}{} 
		&  \multicolumn{1}{|c||}{\textbf{Distortions}} & \textbf{FS} & \multicolumn{1}{c|}{\textbf{Ours}} & \textbf{FS} &  \multicolumn{1}{c|}{\textbf{Ours}}  & \textbf{FS} &  \multicolumn{1}{c|}{\textbf{Ours}}  \\
		
		\cline{2-8} \noalign{\smallskip} \hline
		
		
		\multirow{3}*{\rotatebox[origin=c]{90}{\parbox[c]{1.8cm}{\centering \textbf{CIFAR10}}}}& \multirow{1}*{\textbf{Median}}  & 80.69 & \multicolumn{1}{c|}{93.68} & 80.73 &  \multicolumn{1}{c|}{93.72}  & \textbf{76.15} &  \multicolumn{1}{c|}{69.01} \\
		
		\cline{2-8} 
		
		& \multirow{1}*{\textbf{Bit-depth}} & \textbf{87.96} & \multicolumn{1}{c|}{94.41} & \textbf{86.84} &  \multicolumn{1}{c|}{94.71}  & 52.64 &  \multicolumn{1}{c|}{56.83} \\
		
		\cline{2-8} 
		
		& \multirow{1}*{\textbf{Gray-scale}}  & 75.16 & \multicolumn{1}{c|}{88.85} & 74.62 &  \multicolumn{1}{c|}{88.32}  & 65.09 &  \multicolumn{1}{c|}{65.76} \\
		
		\cline{2-8} 
		
		& \multirow{1}*{\textbf{2 dist.}} & 82.34 & \multicolumn{1}{c|}{96.58} & 82.21 &  \multicolumn{1}{c|}{97.00}  & 74.90 &  \multicolumn{1}{c|}{67.52} \\
		\cline{2-8} 
		
		& \multirow{1}*{\textbf{3 dist.}} & 75.63 & \multicolumn{1}{c|}{\textbf{96.79}} & 75.37 &  \multicolumn{1}{c|}{\textbf{97.08}}  & 73.88 &  \multicolumn{1}{c|}{\textbf{72.50}} \\
		\cline{2-8} 
		
		\hline \noalign{\smallskip} \hline
		
		
		\multirow{3}*{\rotatebox[origin=c]{90}{\parbox[c]{1.8cm}{\centering \textbf{GTSRB}}}}& \multirow{1}*{\textbf{Median}}  & \textbf{97.86} & \multicolumn{1}{c|}{99.51} & \textbf{97.63} &  \multicolumn{1}{c|}{99.46}  & 74.95 &  \multicolumn{1}{c|}{\textbf{75.63}} \\
		
		\cline{2-8} 
		
		& \multirow{1}*{\textbf{Bit-depth}} & 97.00 & \multicolumn{1}{c|}{99.05} & 96.19 &  \multicolumn{1}{c|}{97.36}  & 61.75 &  \multicolumn{1}{c|}{66.95} \\
		
		\cline{2-8} 
		
		& \multirow{1}*{\textbf{Gray-scale}}  & 91.35 & \multicolumn{1}{c|}{96.98} & 90.84 &  \multicolumn{1}{c|}{96.60}  & \textbf{77.19} &  \multicolumn{1}{c|}{73.84} \\
		
		\cline{2-8} 
		
		& \multirow{1}*{\textbf{2 dist.}} & 96.55 & \multicolumn{1}{c|}{99.77} & 96.16 &  \multicolumn{1}{c|}{99.62}  & 73.60 &  \multicolumn{1}{c|}{74.08} \\
		\cline{2-8} 
		
		& \multirow{1}*{\textbf{3 dist.}} & 89.98 & \multicolumn{1}{c|}{\textbf{99.81}} & 89.31 &  \multicolumn{1}{c|}{\textbf{99.72}}  & 76.08 &  \multicolumn{1}{c|}{73.67} \\
		\cline{2-8} 
		
		\hline
		
	\end{tabular}
	\caption{AUC scores of FS and our detector for different choices of the distortions. The configuration with two distortions uses median filtering and bit-depth reduction. All results are reported in percentage (\%).}
	\label{tab_auc_dist}
\end{table}

\subsection{Black-box Results}

The much lower number of samples which succesfully fool the models in black-box settings does not allow us to use the AUC scores as performance quantifiers. Thus, according to \cite{bFS}, we decided to set the threshold such that it rejects only 5\% of the legitimate test samples and collect the detection rates. Table \ref{tab_auc_bb} reports the results. It is clear that our detector outperforms FS even in black-box settings. We believe that the use of per-class statistics provides the main contribution in this setting, forcing the reference vector to be well-shaped, regardless of the effectiveness of the malicious input.

\begin{table}[t!]
	\def\arraystretch{.9}
	\centering
	\scriptsize
	\begin{tabular}{|c||c|c||c|*{3}{c}|}
		\cline{2-7} 
		
		\multicolumn{1}{c}{} \rule{0cm}{7pt} & \multicolumn{1}{|c|}{\textbf{Method}} & \textbf{Threshold} & \textbf{Legitim}. & \textbf{C\&W}  & \textbf{DF} &  \textbf{FGSM4} \\
		
		\cline{2-7} \noalign{\smallskip} \hline
		
		\multirow{4}*{\rotatebox[origin=c]{90}{\parbox[c]{1.1cm}{\centering \textbf{CIFAR10}}}}& \multirow{2}*{\textbf{FS}}  &  \multirow{2}*{1.2119}  &  \multirow{2}*{95.42}  &  \multirow{2}*{7.68}  &  \multirow{2}*{7.58}  &  \multirow{2}*{25.83}    \\
		
		&  & & & & & \\
		
		\cline{2-7} 
		
		& \multirow{2}*{\textbf{Ours}} &  \multirow{2}*{0.6894}  &  \multirow{2}*{94.59}  &  \multirow{2}*{\textbf{22.82}}  &  \multirow{2}*{\textbf{23.06}}  &  \multirow{2}*{\textbf{31.27}}    \\
		
		&  & & & & & \\
		
		\hline \noalign{\smallskip} \hline
		
		
		\multirow{3}*{\rotatebox[origin=c]{90}{\parbox[c]{1.15cm}{\centering \textbf{GTSRB}}}}& \multirow{2}*{\textbf{FS}}  &  \multirow{2}*{0.4097}  &  \multirow{2}*{90.13}  &  \multirow{2}*{69.27}  &  \multirow{2}*{70.15}  &  \multirow{2}*{68.51}   \\
		
		&  & & & & & \\
		
		\cline{2-7} 
		
		& \multirow{2}*{\textbf{Ours}}  &  \multirow{2}*{0.9739}  &  \multirow{2}*{89.59}  &  \multirow{2}*{\textbf{78.28}}  &  \multirow{2}*{\textbf{77.59}}   &  \multirow{2}*{\textbf{72.95}}  \\
		
		&  & & & & & \\
		
		\hline

	\end{tabular}
	\caption{Detection rates of FS and our detector in black-box settings. All results are reported in percentage (\%).}
	 \label{tab_auc_bb}
\end{table}

\section{Conclusions}\label{Concl}

The approach devised results in a cost-effective, reliable detector showing an impressive ability to identify adversarial samples. Compared to FS,our method:
\begin{itemize}
	\item performs better on a broad variety of classifiers. 
	\item is more compatible with other defensive techniques such as adversarial training.
	\item combines in a better way the effects coming from multiple distortions.
	\item performs better in both white and black-box settings.
\end{itemize}  

Despite the experiments and the development presented in this work, there are multiple directions for further improvement.

As suggested in section \ref{CwAT}, it would be interesting to explore the parameter space of the adversarial training defensive technique, finding out the configuration which benefits the most from our detector. Furthermore, it would be intriguing to see how the proposed detector reacts when combined with other defensive strategies.

Another promising work direction consists of merging various distortions to be used as a single one (using less aggressive settings), achieving a distortion able to deal with most of the attack techniques. This would reduce the number of queries to the model, thus reducing the overall inference time.

Finally, we think that the normalized projection score might be used directly as an indicator of the input nature (it is close to $0$ for adversarial samples, whilst close to $1$ for legitimate ones), thus multiplying directly the classification output rather than undergoing a thresholding operation.

\bibliographystyle{aaai}
\fontsize{9.0pt}{10.0pt} \selectfont
\bibliography{bibliography}

\end{document}